%% file: paper.tex
\documentclass[]{fairmeta}
\usepackage{amsmath,amssymb}
\usepackage{array}
\usepackage{booktabs}
\usepackage{makecell}
\usepackage{multirow}
\usepackage{colortbl}
\usepackage[table]{xcolor}
\usepackage{graphicx}
\usepackage{adjustbox}
\usepackage{caption}
\usepackage{float}
\usepackage{bbm}
\usepackage{mathrsfs}
\usepackage{arydshln}
\usepackage{bbding}
\usepackage{xspace}

\newcommand{\algname}{EndoVLM\xspace}
\let\cite\citep

\title{\algname: An Endoscopy Vision-Language Pre-training Model via Anatomy-Guided Sparsity and Progressive Alignment}

\author[*,1,3]{Zhenyu Yi}
\author[*\dagger1,4]{Jianwei Xu}
\author[2,5]{Yue Hu}
\author[1,4]{Zhongwei Qiu}
\author[6]{Sijing Li}
\author[2,5]{Liang Huang}
\author[\dagger2,5]{Bin Lv}
\author[1,4]{Ling Zhang}
\author[1,4]{Yingda Xia}

\affiliation[1]{DAMO Academy, Alibaba Group}
\affiliation[2]{Department of Gastroenterology, The First Affiliated Hospital of Zhejiang Chinese Medical University}
\affiliation[3]{Shanghai Jiao Tong University}
\affiliation[4]{Hupan Lab}
\affiliation[5]{Key Laboratory of Digestive Pathophysiology of Zhejiang Province, The First Affiliated Hospital of Zhejiang Chinese Medical University}
\affiliation[6]{Zhejiang University}

\contribution[*]{Equal contribution}
\contribution[\dagger]{Corresponding author}
\metadata[Code]{\url{https://github.com/Scatteredrain/EndoVLM}}

\input{sec/0_abstract}

\begin{document}
\thispagestyle{firstheader}
\maketitle
\pagestyle{empty}

\input{sec/1_introduction}
\input{sec/2_relatedwork}
\input{sec/3_method}
\input{sec/4_experiment}
\input{sec/5_conclusion}
\input{sec/6_ack}

\bibliographystyle{assets/plainnat}
\bibliography{paper}


\end{document}

%% file: sec/0_abstract.tex
\abstract{
The development of foundation models (FMs) is crucial for advancing endoscopic image analysis. 
However, existing endoscopy FMs mainly rely on self-supervised learning from uni-modal images or videos, overlooking the rich semantic knowledge contained in clinical reports. Furthermore, effectively leveraging these records is hindered by a fundamental modality gap: structured anatomical descriptions are not naturally mapped to specific frames within the high-redundancy, uncurated visual streams.
In this paper, we present EndoVLM, a novel vision-language FM pre-trained on over 348K endoscopic examinations, each pairing a clinical report with its corresponding image collection. 
An Anatomy-Guided Sparse Pooling mechanism utilizes textual descriptions as queries to drive sparse attention, efficiently aggregating semantically salient frames into anatomy-specific visual representations across redundant image-sets. 
Next, a Progressive Semantic-Aware Alignment strategy models clinical taxonomy (anatomy and pathological status) via structured soft targets, bridging the gap from global patient-level matching to fine-grained localized alignment. 
Finally, a Semantic-Concentrated Masked Autoencoder is applied exclusively to these semantic-rich frames, integrating low-level visual precision with robust high-level semantic representation.
Extensive experiments across various downstream tasks demonstrate that EndoVLM outperforms existing foundation models and remains competitive with task-specific methods. 
Remarkably, EndoVLM also exhibits robust zero-shot generalization capabilities, highlighting its potential for broader clinical application. 
}

%% file: sec/1_introduction.tex
\section{Introduction}

Recent advances in foundation models (FMs) have catalyzed a paradigm shift in the development of medical image analysis~\cite{bommasani2021opportunities}. Through self-supervised learning (SSL) on massive unlabeled datasets~\cite{mae,oquab2023dinov2}, FMs learn highly transferable representations for downstream tasks~\cite{xu2021real,hu2024sali,hu2025monobox,limuc}, driving breakthroughs across various medical domains, including CT~\cite{tang2022self}, X-ray~\cite{tiu2022expert}, and histopathology~\cite{wang2024pathology}. Furthermore, since clinical workflows naturally pair images with detailed diagnostic reports, this multimodal synergy has inspired vision-language FMs—such as BiomedCLIP~\cite{zhang2023biomedclip}, LLaVA-Med~\cite{llavamed}, fVLM~\cite{fvlm}, and TumorChain~\cite{li2026tumorchain}. By aligning visual features with rich textual insights, these models build more interpretable and clinically grounded AI systems.

However, existing endoscopy-specific FMs~\cite{endofm,endofmlv,tian2025endomamba,gastronet5m} remain predominantly visual self-supervised models, overlooking the rich clinical insights embedded in textual reports, which not only describe anatomical landmarks and pathological findings but also provide procedural context. Meanwhile, general vision-language pre-training models (VLMs) such as CLIP~\cite{clip,wang2022medclip} typically rely on a one-to-one alignment between individual samples (e.g., a single image, short video clip, or standardized 3D medical data such as CT) and their textual descriptions. 
In contrast, routine gastrointestinal (GI) endoscopy produces a single comprehensive report for dozens of unannotated images or lengthy videos—which we formulate as an \textit{image-set}.
Without explicit temporal or spatial annotations, it is exceptionally difficult to align a specific frame or a local region with a particular sentence or abnormal finding within the comprehensive report~\cite{miech2020end}. 
For example, a gastroscopy report covers \textbf{eight} regions (\textit{esophagus, cardia, fundus, body, antrum, angularis, pylorus, duodenum}), while a colonoscopy assesses \textbf{nine} segments (\textit{ileum, ileocecal region, ascending colon, hepatic flexure, transverse colon, splenic flexure, descending colon, sigmoid colon, rectum}). During examination, clinicians capture multiple images of each region from various angles and document morphology, mucosal features, and any pathological findings.
The lack of per-frame labels makes it nearly impossible to identify the anatomical location or mucosal features in individual images, impeding precise image- or pixel-level cross-modal alignment. Addressing this bottleneck requires novel strategies that bridge structured clinical narratives with large-scale, unstructured visual corpora.

In this work, we present \algname, a novel vision-language foundation model pre-trained on a massive cohort of 348K endoscopic cases. To bridge the gap between unordered image-sets and structured reports, we introduce three core components. First, an Anatomy-Guided Sparse Pooling (AGSP) mechanism employs query-driven sparse attention to distill semantically salient frames from redundant visual streams. Subsequently, we propose a Progressive Semantic-Aware Alignment (PSAA) strategy. By modeling clinical classifications—encompassing both anatomical structures and pathological states—through soft targets, PSAA facilitates a robust transition from coarse patient-level matching to fine-grained localized alignment. Finally, a Semantic-Concentrated Masked Autoencoder (SC-MAE) is applied strictly to these filtered frames, unifying low-level geometric precision with high-level clinical semantics.
Extensive experiments across diverse downstream tasks—including risk stratification, polyp segmentation, anatomy recognition, and video disease diagnosis—demonstrate that \algname outperforms state-of-the-art foundation models and remains competitive with task-specific segmentation baselines using a minimal segmentation head. Remarkably, \algname also exhibits robust zero-shot generalization capabilities, underscoring its immense potential as a highly adaptable backbone for scalable clinical deployment.

%% file: sec/2_relatedwork.tex

%% file: sec/3_method.tex
\section{Method}
\begin{figure}[t]
\centering
\includegraphics[width=\textwidth]{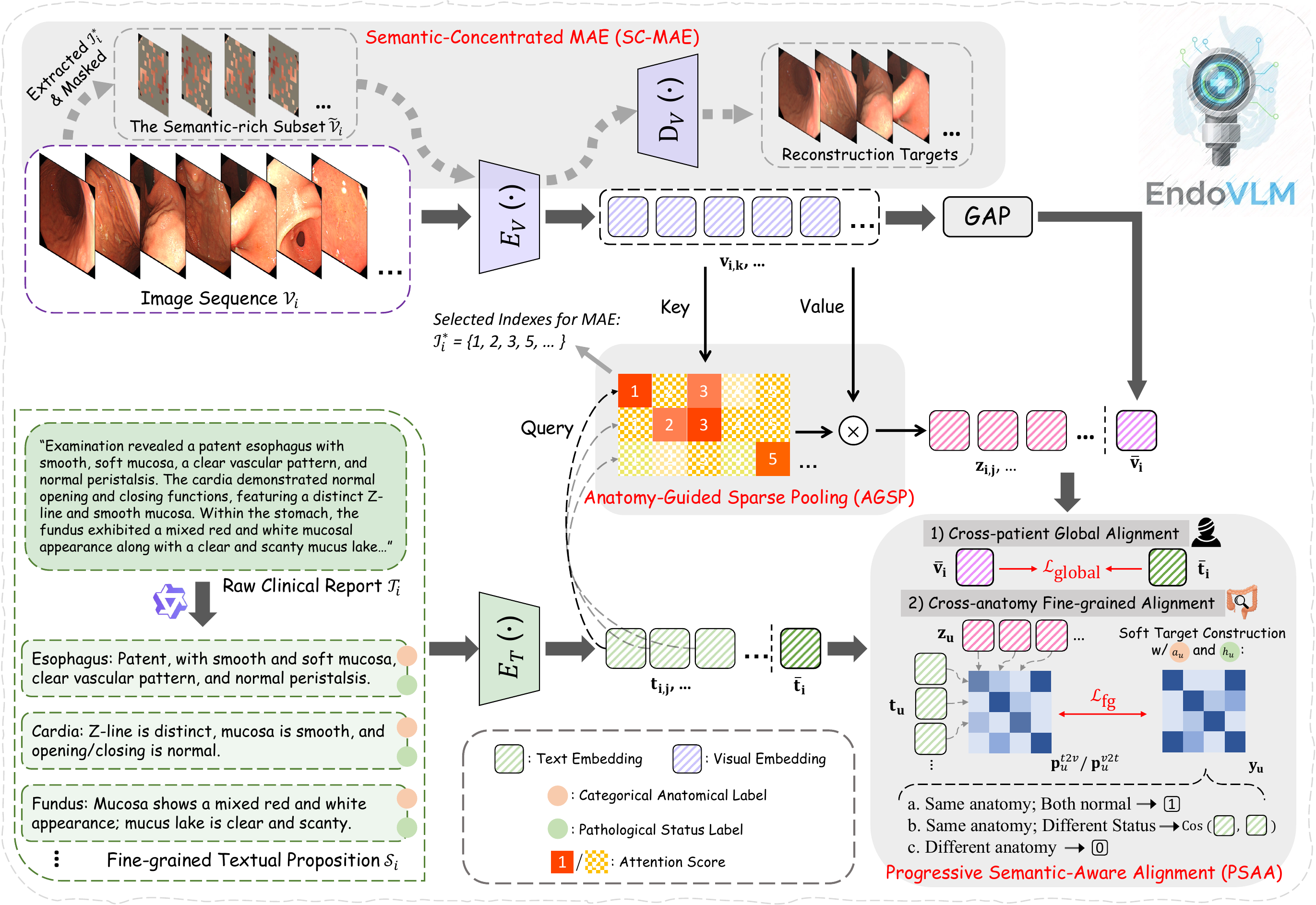}
\caption{\textbf{Overview of \algname.} We align unordered images with reports via:  AGSP for report-guided salient frame distillation and visual feature pooling, PSAA for hierarchical alignment, and SC-MAE for geometric regularization.}
\label{fig:overview}
\end{figure}

We propose \algname, a vision-language foundation model tailored for GI endoscopy. As shown in Fig.~\ref{fig:overview}, to address the asymmetry between high-redundancy, unordered image-sets and raw clinical reports, our approach begins with a data construction pipeline, and followed by three core components: AGSP to distill semantically salient frames via anatomical queries; PSAA for hierarchical semantic alignment; and SC-MAE for robust visual representation learning.

\subsection{Preliminaries and Data Construction}

Given a dataset $\mathcal{D} = \{(\mathcal{V}_i, \mathcal{T}_i)\}_{i=1}^N$ of $N$ examinations, each contains an unordered image-set $\mathcal{V}_i = \{I_{i,k}\}_{k=1}^{K_i}$ and a raw report $\mathcal{T}_i$. Standard global alignment paradigms (e.g., CLIP~\cite{clip}) are suboptimal in this context due to the ``many-to-many'' semantic mismatch between visually redundant frames and information-dense textual findings. To bridge this semantic gap, we use Qwen3~\cite{yang2025qwen3} with a fixed clinical schema to parse $\mathcal{T}_i$ into $M_i$ semantic triplets $\mathcal{S}_i = \{(s_{i,j}, a_{i,j}, h_{i,j})\}_{j=1}^{M_i}$. Here, $s_{i,j}$ represents a fine-grained textual proposition of a distinct morphological finding; $a_{i,j}$ is its anatomical label (mapped to the aforementioned 17 GI regions); and $h_{i,j} \in \{0, 1\}$ indicates the pathological status (normal or abnormal). Detailed prompts, field definitions, and parsing examples are provided in our code repository.

For feature extraction, a vision encoder $E_V$ maps frames $I_{i,k}$ to local normalized embeddings $\mathbf{F}_i^v = \{\mathbf{v}_{i,k} \in \mathbb{R}^d\}_{k=1}^{K_i}$ and a global average-pooled vector $\bar{\mathbf{v}}_i$. Concurrently, a text encoder $E_T$ encodes the full report $\mathcal{T}_i$ and propositions $s_{i,j}$ into a global normalized embedding $\bar{\mathbf{t}}_i$ and a set of fine-grained normalized embeddings $\mathbf{H}_i^t = \{\mathbf{t}_{i,j} \in \mathbb{R}^d\}_{j=1}^{M_i}$, respectively. The final representation for the $i$-th examination is formulated as the tuple: $(\mathbf{F}_i^v, \bar{\mathbf{v}}_i, \bar{\mathbf{t}}_i, \{\mathbf{t}_{i,j}\}, \{a_{i,j}\}, \{h_{i,j}\})$.

\subsection{Anatomy-Guided Sparse Pooling (AGSP)}
Endoscopic image-sets exhibit high redundancy (e.g., repetitive normal mucosa) that dilutes sparse abnormal signals. We propose AGSP, which uses fine-grained text embeddings as queries to selectively aggregate relevant visual evidence.

For each sentence embedding $\mathbf{t}_{i,j}$, we compute dot-product relevance $r_{j,k} = \mathbf{t}_{i,j}^\top \mathbf{v}_{i,k}$ with all frames in $\mathcal{V}_i$. Instead of standard cross attention, we use sparse attention by selecting top-$K$ similar frames $\mathcal{I}_{i,j}$. The anatomy-specific visual representation $\mathbf{z}_{i,j}$ is derived via re-normalized attention over $\mathcal{I}_{i,j}$:
\begin{equation}
    \mathbf{z}_{i,j} = \sum_{k \in \mathcal{I}_{i,j}} \frac{e^{r_{j,k} / \tau_a}}{\sum_{m \in \mathcal{I}_{i,j}} e^{r_{j,m} / \tau_a}} \mathbf{v}_{i,k},
\end{equation}
where $\tau_a = 0.07$ is the temperature and $K = 3$. A small $K$ preserves a compact set of salient frames for each anatomical query while suppressing redundant normal views. This aligns the visual modality with the anatomy described in $\mathbf{t}_{i,j}$ while filtering noise.

\subsection{Progressive Semantic-Aware Alignment (PSAA)}
To capture both holistic context and fine-grained mucosal and pathological nuances, we employ a progressive alignment strategy.

\noindent\textbf{Stage 1: Cross-Patient Global Alignment.} 
We first enforce consistency between the patient-level image-set representation $\bar{\mathbf{v}}_i$ and the global report embedding $\bar{\mathbf{t}}_i$ using the symmetric InfoNCE~\cite{infonce} loss across a batch of size $B$:
\begin{equation}
    \mathcal{L}_{\text{global}} = - \frac{1}{2B} \sum_{i=1}^B \left( \log \frac{e^{\langle \bar{\mathbf{v}}_i, \bar{\mathbf{t}}_i \rangle / \tau}}{\sum_{j=1}^{B} e^{\langle \bar{\mathbf{v}}_i, \bar{\mathbf{t}}_j \rangle / \tau}} + \log \frac{e^{\langle \bar{\mathbf{t}}_i, \bar{\mathbf{v}}_i \rangle / \tau}}{\sum_{j=1}^{B} e^{\langle \bar{\mathbf{t}}_i, \bar{\mathbf{v}}_j \rangle / \tau}} \right),
\end{equation}
where $\langle \cdot, \cdot \rangle$ is cosine similarity and $\tau$ is a learnable temperature.

\noindent\textbf{Stage 2: Cross-Anatomy Fine-Grained Alignment.} 
Global alignment alone is insufficient for distinguishing subtle mucosa attributes. Inspired by recent dense semantic alignment formulations~\cite{fvlm,wang2025simcrop}, we perform fine-grained contrastive learning on a set of semantic units $\mathcal{U} = \{(\mathbf{z}_u, \mathbf{t}_u)\}_{u=1}^{M}$, aggregating all $M$ valid local visual-text pairs $(\mathbf{z}_{i,j},\mathbf{t}_{i,j})$ in the mini-batch. 

We first compute the softmax-normalized image-to-text and text-to-image similarity distributions, denoted as $\mathbf{p}_u^{\text{v2t}}$ and $\mathbf{p}_u^{\text{t2v}}$, respectively. For any pair of units $(u, k)$ within $\mathcal{U}$, the $k$-th element of these distributions is defined as:
\begin{equation}
    p_{u,k}^{\text{v2t}} = \frac{\exp(\langle \mathbf{z}_u, \mathbf{t}_k \rangle / \tau)}{\sum_{m=1}^{M} \exp(\langle \mathbf{z}_u, \mathbf{t}_m \rangle / \tau)}, \quad 
    p_{u,k}^{\text{t2v}} = \frac{\exp(\langle \mathbf{t}_u, \mathbf{z}_k \rangle / \tau)}{\sum_{m=1}^{M} \exp(\langle \mathbf{t}_u, \mathbf{z}_m \rangle / \tau)}.
\end{equation}

To seamlessly progress from global cross-patient matching to localized alignment of anatomy and pathological status, we construct a taxonomy-aware soft target distribution $\mathbf{y}_u = [y_{u,1}, \dots, y_{u,M}]^\top \in \mathbb{R}^M$. Let $a_u$ and $h_u$ be the categorical anatomical and pathological status labels of the $u$-th unit, respectively. The normalized target $y_{u,k}$ is computed as:
\begin{equation}
    y_{u,k} = \frac{\hat{y}_{u,k}}{\sum_{m=1}^{M} \hat{y}_{u,m}}, \text{ with } \hat{y}_{u,k} = 
    \begin{cases} 
    1, & \text{if } a_u = a_k \text{ and } h_u = h_k = 0; \\
    \langle \mathbf{t}_u, \mathbf{t}_k \rangle, & \text{if } a_u = a_k \text{ and } (h_u=1 \text{ or } h_k=1); \\
    0, & \text{otherwise.}
    \end{cases}
\end{equation}

By avoiding repulsion among identical healthy anatomies and adaptively scaling abnormal alignments via textual similarity, this formulation drives the model to autonomously disentangle anatomical and pathological semantics. The textual similarity term provides a semantic relational prior within each anatomical group, allowing related abnormal findings to share softer supervision. The fine-grained alignment loss is defined as follows, where $\text{CE}$ denotes cross-entropy.
\begin{equation}
    \mathcal{L}_{\text{fg}} = \frac{1}{2M} \sum_{u=1}^{M} \left( \text{CE}(\mathbf{y}_u, \mathbf{p}_u^{v2t}) + \text{CE}(\mathbf{y}_u, \mathbf{p}_u^{t2v}) \right).
\end{equation}

\subsection{Semantic-Concentrated Masked Autoencoder (SC-MAE)}
To efficiently capture low-level textures, SC-MAE applies the MAE paradigm~\cite{mae} exclusively to the semantic-rich subset $\tilde{\mathcal{V}}_i = \{ I_{i,k} \in \mathcal{V}_i \mid k \in \mathcal{I}_i^* \}$, where $\mathcal{I}_i^* = \bigcup_{j} \mathcal{I}_{i,j}$ denotes the unified index set aggregated from all anatomy-specific frames retrieved by AGSP. It masks $75\%$ of patches in $\tilde{\mathcal{V}}_i$, and reconstruct pixels via a decoder $D_V$, yielding an MSE loss $\mathcal{L}_{\text{MAE}}$. The overall training objective is as follows, where $\lambda_1, \lambda_2$ are balancing weights.
\begin{equation}
    \mathcal{L} = \mathcal{L}_{\text{global}} + \lambda_1 \mathcal{L}_{\text{fg}} + \lambda_2 \mathcal{L}_{\text{MAE}}.
\end{equation}

%% file: sec/4_experiment.tex
\section{Experiments and Results}
\begin{table}[t]
\centering
\caption{Performance comparison. We report F1 (\%) for PolypDiag, macro AUC/F1 (\%) for LIMUC, Dice (\%) for polyp segmentation (CVC-12k, Kvasir (Kva), ClinicDB (Clinic), Endoscene (Endo), ColonDB, ETIS). $^{\dagger}$: pre-trained with our dataset. '--': not reported and no public code or weights. The best and second-best results are in bold and underlined.}
\label{tab:comprehensive_results}
\small

\renewcommand{\arraystretch}{1.05}

\setlength{\tabcolsep}{2.0pt}
\begin{tabular*}{\textwidth}{@{\extracolsep{\fill}}>{\raggedright\arraybackslash}p{5.35cm} cc c ccccc@{}}
\toprule
\multirow{2}{*}{\makecell[c]{\textbf{Model}}} & \multirow{2}{*}{\makecell[c]{\textbf{Polyp}\\\textbf{Diag}}} & \multirow{2}{*}{\makecell[c]{\textbf{LIMUC}}} & \multirow{2}{*}{\makecell[c]{\textbf{CVC}\\\textbf{-12k}}} & \multicolumn{2}{c}{\textbf{Seen}} & \multicolumn{3}{c}{\textbf{Unseen}} \\
\cmidrule(lr){5-6} \cmidrule(lr){7-9}
 &  &  &  & \textbf{Kva} & \textbf{Clinic} & \textbf{Endo} & \textbf{Colon} & \textbf{ETIS} \\ 
\midrule
\rowcolor{gray!15} \multicolumn{9}{l}{\textit{General FMs}} \\
MAE~\cite{mae}                       & 91.1 & 93.4/72.7 & 83.6 & 88.4 & 88.5 & 84.0 & 75.5 & 74.7 \\
CLIP~\cite{clip}                     & 90.0 & 92.0/69.2 & 80.0 & 88.3 & 90.0 & 85.4 & 75.8 & 71.6 \\
DINOv2~\cite{oquab2023dinov2}        & 93.3 & 93.8/73.6 & 83.8 & 90.7 & 91.4 & 88.6 & 79.8 & 77.8 \\
DINOv3~\cite{simeoni2025dinov3}      & 93.9 & 93.9/73.7 & 84.5 & 91.0 & 91.1 & 89.1 & \underline{81.6} & 77.7 \\
BiomedCLIP~\cite{zhang2023biomedclip}& 90.2 & 92.6/72.0 & 81.5 & 88.0 & 87.8 & 82.8 & 72.7 & 66.1 \\
{MAE$^{\dagger}$}~\cite{mae}           & 93.7 & 93.1/73.6 & 84.2 & 88.4 & 88.5 & 83.9 & 75.5 & 74.7 \\
{CLIP$^{\dagger}$}~\cite{clip}     & 90.1 & 85.2/60.2 & 66.2 & 60.1 & 70.6 & 28.8 & 32.8 & 25.6 \\
{DINOv3$^{\dagger}$}~\cite{simeoni2025dinov3}& 94.8 & 94.0/\underline{74.0} & \underline{85.8} & 91.4 & 92.2 & 89.5 & \underline{81.6} & 80.1 \\
\midrule
\rowcolor{gray!15} \multicolumn{9}{l}{\textit{Endo SSL}} \\
EndoFM~\cite{endofm}                 & 90.7 & 93.0/72.9 & 73.9 & 87.7 & 87.1 & 83.2 & 71.2 & 63.5 \\
EndoFM-LV~\cite{endofmlv}            & \underline{96.3} & 83.3/57.4 & 83.2 & 62.2 & 67.7 & 42.5 & 35.0 & 29.8 \\
EndoMamba~\cite{tian2025endomamba}   & 95.0 & 92.1/70.1 & 85.4 & 87.5 & 87.8 & 86.2 & 69.4 & 60.2 \\
EndoDINO~\cite{dermyer2025endodino}  & --   & 93.7/70.6 & --   & --   & --   & --   & --   & --   \\
GastroNet-5M~\cite{gastronet5m}      & --   & \multicolumn{1}{l}{\textbf{95.5}/--} & --   & --   & --   & --   & --   & --   \\
\midrule
\rowcolor{gray!15} \multicolumn{9}{l}{\textit{Task specific model (Segmentation)}} \\
Polyp-PVT~\cite{polyppvt}            & --   & --/--     & --   & \underline{91.7}  & \underline{93.7} & 90.0  & 80.8 & 78.7 \\
VM-UNet~\cite{vmunet}                & --   & --/--     & --   & 91.3 & 92.6 & 88.6 & 79.8 & 76.1 \\
PolyMamba~\cite{polymamba}           & --   & --/--     & --   & \textbf{91.9} & \textbf{94.0} & \underline{90.4} & {81.5} & \textbf{82.9} \\
\midrule
\textbf{EndoVLM}              & \textbf{97.3} & \underline{94.5}/\textbf{74.4} & \textbf{86.4} & \textbf{91.9} & 93.1 & \textbf{90.8} & \textbf{82.8} & \underline{82.0} \\
\bottomrule
\end{tabular*}
\end{table}
\noindent\textbf{Pre-training dataset.} We retrospectively collected over 400K endoscopic examinations from two medical centers\footnote{The study was conducted in accordance with the Declaration of Helsinki and approved by the institutional review board of the lead medical center (No. 2024-KLS-489-02).}, each consisting of a clinical report and its corresponding image package. By leveraging Qwen3~\cite{yang2025qwen3} to analyze report content, we filtered out ineligible cases—including post-operative exams, incomplete procedures (defined as gastroscopy reports missing any of the 8 standard anatomical regions, or colonoscopy reports missing any of the 9 intestinal landmarks), and other non-conforming studies. The final curated dataset contains 348K examinations, comprising more than 18.6M endoscopic images.

\noindent\textbf{Downstream task setup.} 
To evaluate the versatility and robustness of EndoVLM, we conducted experiments across various tasks: (i) 
\textit{PolypDiag for video polyp diagnosis}~\cite{polypdiag}. Following EndoFM-LV~\cite{endofmlv}, we sampled 16 frames per video. Frame-level features extracted by our image-based backbone were aggregated via mean pooling to form a video-level representation, which was then classified using a linear head. (ii) 
\textit{CVC-12k for video polyp segmentation}~\cite{ClinicDB} and (iii) \textit{Kvasir-SEG}~\cite{kvasir-seg}, \textit{ClinicDB}~\cite{ClinicDB}, \textit{ColonDB}~\cite{ColonDB}, \textit{ETIS}~\cite{ETIS}, \textit{EndoScene}~\cite{EndoScene} \textit{for polyp segmentation generalization assessment}. A linear segmentation head is implemented by reshaping the backbone's patch tokens into spatial feature maps $(B, D, H/P, W/P)$, followed by \textbf{a single convolutional layer and bilinear interpolation for final dense prediction}. We adopt the same data splits and evaluation protocol as EndoFM-LV~\cite{endofmlv} on CVC-12k, and strictly follow the training and testing setup of Polyp-PVT~\cite{polyppvt} for generalization assessment. For this Polyp-PVT generalization setup, ``seen'' and ``unseen'' denote whether a public segmentation dataset is used during downstream fine-tuning, while all downstream datasets are disjoint from the private pre-training corpus. (iv) 
\textit{LIMUC for ulcerative colitis severity grading}~\cite{limuc}. We append a linear classification head to the EndoVLM, and follow the same experimental settings as GastroNet5M~\cite{gastronet5m}. (v)
\textit{Zero-Shot Transfer.} We assess the model's generalizability via zero-shot anatomical recognition on Hyper-Kvasir~\cite{borgli2020hyperkvasir} (covering upper \& lower GI tracts) and video disease diagnosis on PolypDiag~\cite{polypdiag}, directly leveraging the pre-aligned visual-semantic space without task-specific fine-tuning.

\noindent\textbf{Implementation details.}
The vision encoder and language encoder are instantiated as ViT-B/16~\cite{vit} and PubMedBERT~\cite{pubmedbert}, respectively. 
All input images are resized to $224 \times 224$ for pre-training. The entire framework is trained for 100 epochs on NVIDIA A800 GPUs with a total batch size of 96. We optimize the network using the AdamW optimizer with a base learning rate of 1.5e-4 and a weight decay of 0.05. The loss weights $\lambda_1$ and $\lambda_2$ are empirically set to 1.

\noindent\textbf{Comparison experiment.}
We compare our method against recent SOTA methods, including general visual and vision-language FMs (DINOv2/v3~\cite{oquab2023dinov2,simeoni2025dinov3}, MAE~\cite{mae}, CLIP~\cite{clip}, BiomedCLIP~\cite{zhang2023biomedclip}), endoscopy-specific visual SSL models (EndoFM~\cite{endofm}, EndoFM-LV~\cite{endofmlv}, EndoMamba~\cite{tian2025endomamba}, EndoDINO~\cite{dermyer2025endodino}, GastroNet-5M~\cite{gastronet5m}), and several task-specific models~\cite{polyppvt,polymamba,vmunet} for polyp segmentation. We also pre-train the general FMs from scratch on our dataset. Our approach surpasses all FMs, trailing only GastroNet-5M on LIMUC. Notably, (1) On video tasks (PolypDiag/CVC-12k), our model exceeds video-specific FMs~\cite{endofm,endofmlv,tian2025endomamba} using a simple image-level mean-pooling aggregation, bypassing complex temporal modeling.
(2) For dense prediction tasks (polyp segmentation), we achieve competitive or superior performance using only a single convolutional layer with bilinear interpolation, without the need for complex decoders present in task-specific models~\cite{polyppvt,vmunet,polymamba}. This confirms that our pre-training yields high-quality dense features. Superior results on unseen ColonDB and ETIS datasets further validate our model's robust generalization.

\begin{figure}[t]
    \centering
    \includegraphics[width=0.8\textwidth]{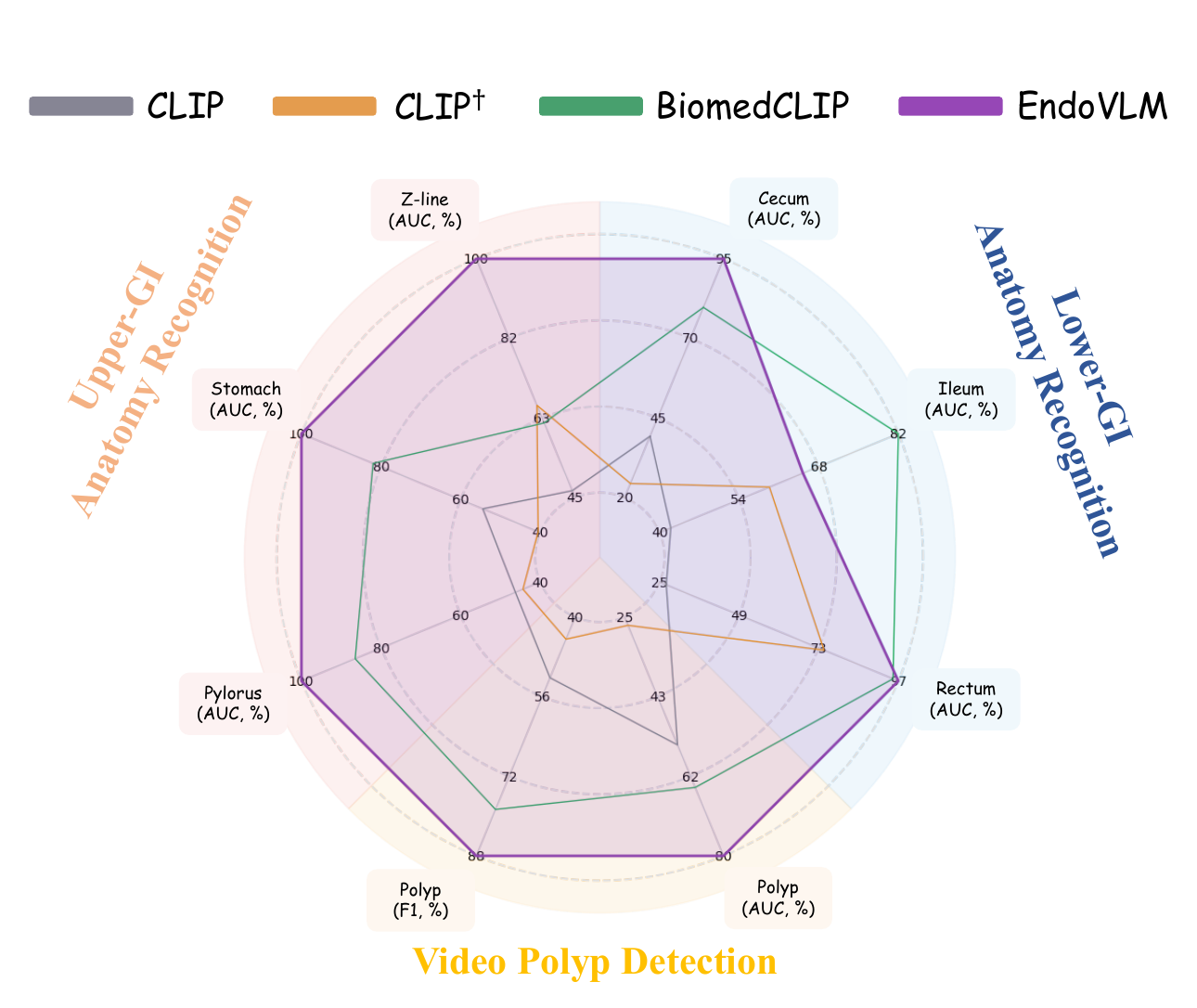}
    \caption{Zero-shot performance comparison.}
    \label{fig:zeroshot}
\end{figure}

\noindent\textbf{Zero-shot evaluation.}
We evaluate zero-shot performance on three clinical classification tasks using the prompt: \textit{"This is an image of \{cls\}."} As shown in Fig.~\ref{fig:zeroshot}, \algname significantly outperforms general (CLIP) and medical-specific (CLIP$^{\dagger}$, BiomedCLIP) FMs. It achieves near-perfect transferability ($\sim100\%$ AUC) in Upper-GI anatomy recognition and surpasses BiomedCLIP by 18\% AUC in the challenging video disease diagnosis task. These results validate that our taxonomy-aware alignment yields robust, highly transferable representations without task-specific fine-tuning. The lower ileum result reflects a category-specific limitation, as ileum descriptions are often procedural and less uniform in colonoscopy reports.

\begin{table}[t]
    \centering
    \caption{Ablation study of different loss components. $\mathcal{L}_{\text{glo}}$: Global Alignment; $\mathcal{L}_{\text{fg}}$: Fine-Grained Alignment; $\mathcal{L}_{\text{MAE}}$: Masked Autoencoder.}
    \label{tab:ablation}
    \footnotesize
    \setlength{\tabcolsep}{14pt}
    \begin{tabular}{ccccc}
        \toprule
        \multicolumn{3}{c}{\textbf{Components}} & \textbf{LIMUC} & \textbf{CVC-12k} \\
        \cmidrule(lr){1-3} \cmidrule(lr){4-5}
        $\mathcal{L}_{\text{glo}}$ & $\mathcal{L}_{\text{fg}}$ & $\mathcal{L}_{\text{MAE}}$ & \textbf{AUC} & \textbf{Dice} \\
        \midrule
        \checkmark & & & 85.2 & 66.2 \\
        \checkmark & \checkmark & & 94.1 & 85.6 \\
        & & \checkmark & 93.1 & 84.2 \\
        \checkmark & \checkmark & \checkmark & \textbf{94.5} & \textbf{86.4} \\
        \bottomrule
    \end{tabular}
\end{table}

\noindent\textbf{Ablation study.}
Ablation results in Table~\ref{tab:ablation} validate the efficacy of our proposed components. Global alignment ($\mathcal{L}_{\text{glo}}$) alone--mimicking standard CLIP paradigm--yields poor performance (85.2\% AUC and 66.2\% Dice), as coarse patient-level matching struggles to capture localized lesion semantics. Integrating fine-grained alignment ($\mathcal{L}_{\text{fg}}$) triggers a significant leap, boosting AUC by 8.9\% and Dice by 19.4\%. which highlights the critical role of our taxonomy-aware soft targets in aligning detailed anatomical and pathological features. Finally, SC-MAE ($\mathcal{L}_{\text{MAE}}$) further refines representations to achieve the best performance (94.5\% AUC and 86.4\% Dice), demonstrating that pixel-level reconstruction provides complementary geometric regularization to semantic alignment. 

\begin{figure}[H]
\centering
\includegraphics[width=1\textwidth]{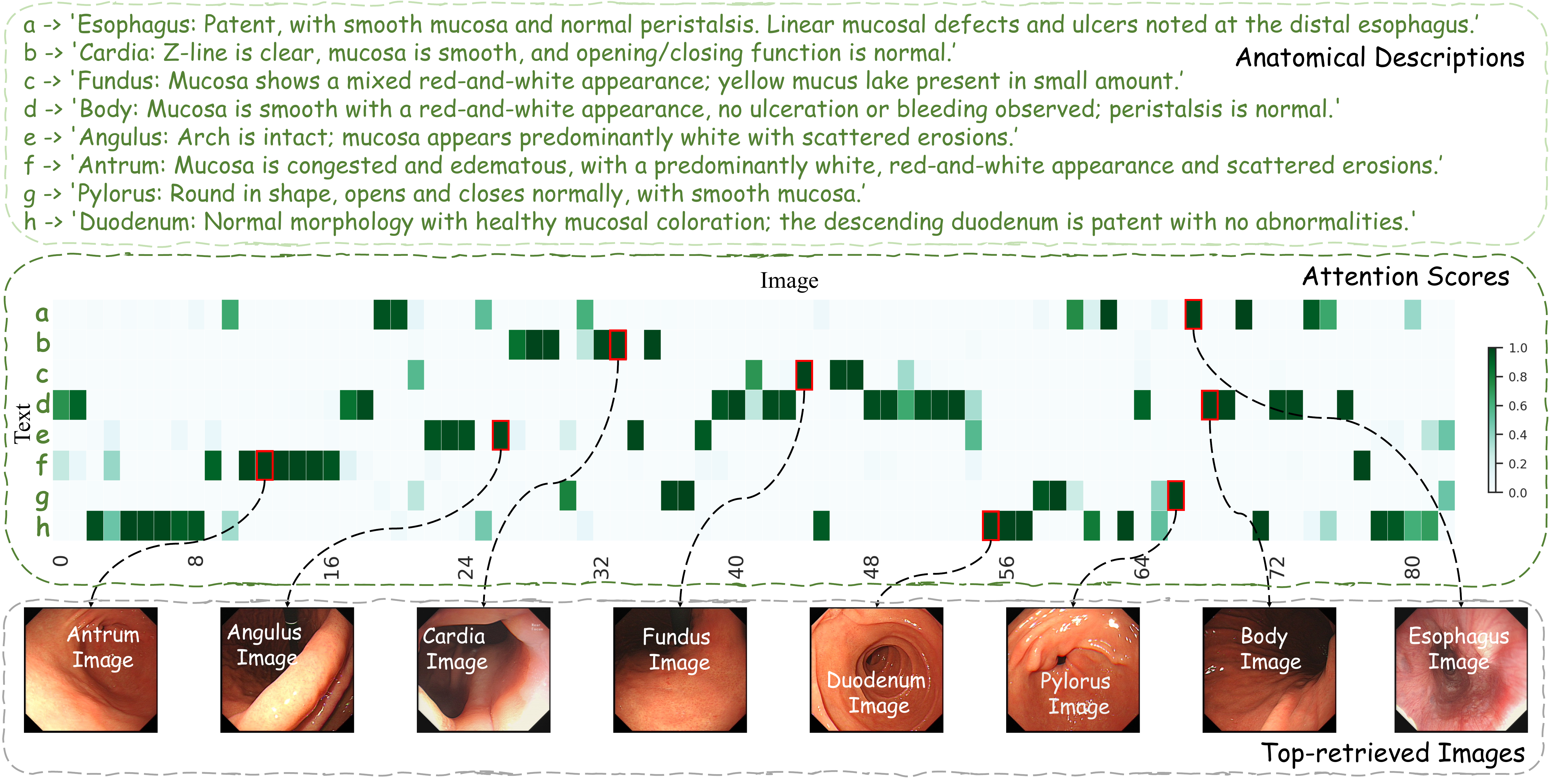}
\caption{Visualization of AGSP \textbf{attention scores}. The heatmap illustrates similarities between unordered image embeddings and anatomical text embeddings, where $a$-$h$ denote specific \textbf{anatomical descriptions} in a gastroscopy report. Highlights identify the \textbf{top-retrieved images} for each specific anatomy.}
\label{fig:agsp_vis}
\end{figure}

\noindent\textbf{Qualitative analysis and interpretability.}
To validate the interpretability of our AGSP module, we visualize the attention scores between the sentence embeddings $\mathbf{t}_{i,j}$ and image embeddings $\mathbf{v}_{i,k}$. As shown in Fig.~\ref{fig:agsp_vis}, the attention matrix exhibits highly localized and target-specific activations. \algname accurately retrieves the most representative frames for eight distinct upper-GI regions (e.g., Esophagus, Duodenum) from a redundant set of over 80 images.

%% file: sec/5_conclusion.tex
\section{Conclusion}
We present EndoVLM, the first GI VLM pre-trained on a large-scale dataset of unordered GI endoscopy image-sets and clinical reports. To bridge the profound semantic gap between redundant visual frames and structured narratives, we propose a unified framework that seamlessly integrates semantics-driven salient frame distillation, progressive cross-modal alignment, and geometric reconstruction. Pre-trained on a massive cohort of 348K cases, EndoVLM demonstrates exceptional adaptability and robust zero-shot generalization. It significantly outperforms existing uni-modal foundation models and complex task-specific architectures across diverse downstream tasks, providing a highly scalable foundation for next-generation AI-assisted endoscopy.

%% file: sec/6_ack.tex
\section*{Acknowledgments}
This study was supported by ``Pioneer'' and ``Leading Goose'' R\&D Program of Zhejiang (2023C03050).

%% file: paper.bbl
\begin{thebibliography}{38}
\providecommand{\natexlab}[1]{#1}
\providecommand{\url}[1]{\texttt{#1}}
\expandafter\ifx\csname urlstyle\endcsname\relax
  \providecommand{\doi}[1]{doi: #1}\else
  \providecommand{\doi}{doi: \begingroup \urlstyle{rm}\Url}\fi

\bibitem[Bernal et~al.(2015)Bernal, S{\'a}nchez, Fern{\'a}ndez-Esparrach, Gil,
  Rodr{\'\i}guez, and Vilari{\~n}o]{ClinicDB}
Jorge Bernal, F~Javier S{\'a}nchez, Gloria Fern{\'a}ndez-Esparrach, Debora Gil,
  Cristina Rodr{\'\i}guez, and Fernando Vilari{\~n}o.
\newblock Wm-dova maps for accurate polyp highlighting in colonoscopy:
  Validation vs. saliency maps from physicians.
\newblock \emph{Computerized medical imaging and graphics}, 43:\penalty0
  99--111, 2015.

\bibitem[Bommasani et~al.(2021)Bommasani, Hudson, Adeli, Altman, Arora, von
  Arx, Bernstein, Bohg, Bosselut, Brunskill,
  et~al.]{bommasani2021opportunities}
Rishi Bommasani, Drew~A Hudson, Ehsan Adeli, Russ Altman, Simran Arora, Sydney
  von Arx, Michael~S Bernstein, Jeannette Bohg, Antoine Bosselut, Emma
  Brunskill, et~al.
\newblock On the opportunities and risks of foundation models.
\newblock \emph{arXiv preprint arXiv:2108.07258}, 2021.

\bibitem[Borgli et~al.(2020)Borgli, Thambawita, Smedsrud,
  et~al.]{borgli2020hyperkvasir}
Hanna Borgli, Vajira Thambawita, Pia~H Smedsrud, et~al.
\newblock Hyperkvasir, a comprehensive multi-class image and video dataset for
  gastrointestinal endoscopy.
\newblock \emph{Scientific data}, 7\penalty0 (1):\penalty0 283, 2020.

\bibitem[Dermyer et~al.(2025)Dermyer, Kalra, and Schwartz]{dermyer2025endodino}
Patrick Dermyer, Angad Kalra, and Matt Schwartz.
\newblock Endodino: A foundation model for gi endoscopy.
\newblock \emph{arXiv preprint arXiv:2501.05488}, 2025.

\bibitem[Dong et~al.(2021)Dong, Wang, Fan, Li, Fu, and Shao]{polyppvt}
Bo~Dong, Wenhai Wang, Deng-Ping Fan, Jinpeng Li, Huazhu Fu, and Ling Shao.
\newblock Polyp-pvt: Polyp segmentation with pyramid vision transformers.
\newblock \emph{arXiv preprint arXiv:2108.06932}, 2021.

\bibitem[Dosovitskiy et~al.(2020)Dosovitskiy, Beyer, Kolesnikov, et~al.]{vit}
Alexey Dosovitskiy, Lucas Beyer, Alexander Kolesnikov, et~al.
\newblock An image is worth 16x16 words: Transformers for image recognition at
  scale.
\newblock \emph{arXiv preprint arXiv:2010.11929}, 2020.

\bibitem[Fu et~al.(2025)Fu, Hu, Zheng, Tang, and Liu]{polymamba}
Renyu Fu, Shurui Hu, Xiao Zheng, Chang Tang, and Xinwang Liu.
\newblock Polymamba: Spatial-prior guided mamba for polyp segmentation with
  high-frequency enhancement.
\newblock In \emph{MICCAI}, pages 455--465. Springer, 2025.

\bibitem[Gu et~al.(2021)Gu, Tinn, Cheng, et~al.]{pubmedbert}
Yu~Gu, Robert Tinn, Hao Cheng, et~al.
\newblock Domain-specific language model pretraining for biomedical natural
  language processing.
\newblock \emph{ACM Transactions on Computing for Healthcare}, 3\penalty0
  (1):\penalty0 1--23, 2021.

\bibitem[He et~al.(2022)He, Chen, Xie, Li, Doll{\'a}r, and Girshick]{mae}
Kaiming He, Xinlei Chen, Saining Xie, Yanghao Li, Piotr Doll{\'a}r, and Ross
  Girshick.
\newblock Masked autoencoders are scalable vision learners.
\newblock In \emph{CVPR}, pages 16000--16009, 2022.

\bibitem[Hu et~al.(2024)Hu, Yi, Zhou, Peng, Liu, Li, and Wang]{hu2024sali}
Qiang Hu, Zhenyu Yi, Ying Zhou, Fang Peng, Mei Liu, Qiang Li, and Zhiwei Wang.
\newblock Sali: Short-term alignment and long-term interaction network for
  colonoscopy video polyp segmentation.
\newblock In \emph{International Conference on Medical Image Computing and
  Computer-Assisted Intervention}, pages 531--541. Springer, 2024.

\bibitem[Hu et~al.(2025)Hu, Yi, Zhou, Huang, Liu, Li, and Wang]{hu2025monobox}
Qiang Hu, Zhenyu Yi, Ying Zhou, Fan Huang, Mei Liu, Qiang Li, and Zhiwei Wang.
\newblock Monobox: Tightness-free box-supervised polyp segmentation using
  monotonicity constraint.
\newblock In \emph{Proceedings of the AAAI Conference on Artificial
  Intelligence}, volume~39, pages 3572--3580, 2025.

\bibitem[Jha et~al.(2019)Jha, Smedsrud, Riegler, Halvorsen, De~Lange, Johansen,
  and Johansen]{kvasir-seg}
Debesh Jha, Pia~H Smedsrud, Michael~A Riegler, P{\aa}l Halvorsen, Thomas
  De~Lange, Dag Johansen, and H{\aa}vard~D Johansen.
\newblock Kvasir-seg: A segmented polyp dataset.
\newblock In \emph{International conference on multimedia modeling}, pages
  451--462. Springer, 2019.

\bibitem[Jong et~al.(2025)Jong, Boers, Fockens, et~al.]{gastronet5m}
Martijn~R Jong, Tim~GW Boers, Kiki~N Fockens, et~al.
\newblock Gastronet-5m: A multicenter dataset for developing foundation models
  in gastrointestinal endoscopy.
\newblock \emph{Gastroenterology}, 2025.

\bibitem[Li et~al.(2023)Li, Wong, Zhang, Usuyama, Liu, Yang, Naumann, Poon, and
  Gao]{llavamed}
Chunyuan Li, Cliff Wong, Sheng Zhang, Naoto Usuyama, Haotian Liu, Jianwei Yang,
  Tristan Naumann, Hoifung Poon, and Jianfeng Gao.
\newblock Llava-med: Training a large language-and-vision assistant for
  biomedicine in one day.
\newblock \emph{NeurIPS}, 36:\penalty0 28541--28564, 2023.

\bibitem[Li et~al.(2026)Li, Qiu, Liu, Zhang, Lin, Xie, An, Yun, Yang, Xiao,
  et~al.]{li2026tumorchain}
Sijing Li, Zhongwei Qiu, Jiang Liu, Wenqiao Zhang, Tianwei Lin, Yihan Xie,
  Jianxiang An, Boxiang Yun, Chenglin Yang, Jun Xiao, et~al.
\newblock Tumorchain: Interleaved multimodal chain-of-thought reasoning for
  traceable clinical tumor analysis.
\newblock In \emph{International Conference on Learning Representations},
  volume 2026, pages 119790--119819, 2026.

\bibitem[Miech et~al.(2020)Miech, Alayrac, Smaira, Laptev, Sivic, and
  Zisserman]{miech2020end}
Antoine Miech, Jean-Baptiste Alayrac, Lucas Smaira, Ivan Laptev, Josef Sivic,
  and Andrew Zisserman.
\newblock End-to-end learning of visual representations from uncurated
  instructional videos.
\newblock In \emph{CVPR}, pages 9879--9889, 2020.

\bibitem[Oord et~al.(2018)Oord, Li, and Vinyals]{infonce}
Aaron van~den Oord, Yazhe Li, and Oriol Vinyals.
\newblock Representation learning with contrastive predictive coding.
\newblock \emph{arXiv preprint arXiv:1807.03748}, 2018.

\bibitem[Oquab et~al.(2023)Oquab, Darcet, Moutakanni, et~al.]{oquab2023dinov2}
Maxime Oquab, Timoth{\'e}e Darcet, Th{\'e}o Moutakanni, et~al.
\newblock Dinov2: Learning robust visual features without supervision.
\newblock \emph{arXiv preprint arXiv:2304.07193}, 2023.

\bibitem[Polat et~al.(2023)Polat, Kani, Ergenc, Ozen~Alahdab, Temizel, and
  Atug]{limuc}
Gorkem Polat, Haluk~Tarik Kani, Ilkay Ergenc, Yesim Ozen~Alahdab, Alptekin
  Temizel, and Ozlen Atug.
\newblock Improving the computer-aided estimation of ulcerative colitis
  severity according to mayo endoscopic score by using regression-based deep
  learning.
\newblock \emph{Inflammatory Bowel Diseases}, 29\penalty0 (9):\penalty0
  1431--1439, 2023.

\bibitem[Radford et~al.(2021)Radford, Kim, Hallacy, et~al.]{clip}
Alec Radford, Jong~Wook Kim, Chris Hallacy, et~al.
\newblock Learning transferable visual models from natural language
  supervision.
\newblock In \emph{ICML}, pages 8748--8763. PmLR, 2021.

\bibitem[Ruan et~al.(2024)Ruan, Li, and Xiang]{vmunet}
Jiacheng Ruan, Jincheng Li, and Suncheng Xiang.
\newblock Vm-unet: Vision mamba unet for medical image segmentation.
\newblock \emph{ACM Transactions on Multimedia Computing, Communications and
  Applications}, 2024.

\bibitem[Shui et~al.(2025)Shui, Zhang, Cao, Wang, Guo, Lu, Yang, Ye, Liang,
  Zhang, et~al.]{fvlm}
Zhongyi Shui, Jianpeng Zhang, Weiwei Cao, Sinuo Wang, Ruizhe Guo, Le~Lu, Lin
  Yang, Xianghua Ye, Tingbo Liang, Qi~Zhang, et~al.
\newblock Large-scale and fine-grained vision-language pre-training for
  enhanced ct image understanding.
\newblock \emph{arXiv preprint arXiv:2501.14548}, 2025.

\bibitem[Silva et~al.(2014)Silva, Histace, Romain, Dray, and Granado]{ETIS}
Juan Silva, Aymeric Histace, Olivier Romain, Xavier Dray, and Bertrand Granado.
\newblock Toward embedded detection of polyps in wce images for early diagnosis
  of colorectal cancer.
\newblock \emph{International journal of computer assisted radiology and
  surgery}, 9\penalty0 (2):\penalty0 283--293, 2014.

\bibitem[Sim{\'e}oni et~al.(2025)Sim{\'e}oni, Vo, Seitzer, Baldassarre, Oquab,
  Jose, Khalidov, Szafraniec, Yi, Ramamonjisoa, et~al.]{simeoni2025dinov3}
Oriane Sim{\'e}oni, Huy~V Vo, Maximilian Seitzer, Federico Baldassarre, Maxime
  Oquab, Cijo Jose, Vasil Khalidov, Marc Szafraniec, Seungeun Yi, Micha{\"e}l
  Ramamonjisoa, et~al.
\newblock Dinov3.
\newblock \emph{arXiv preprint arXiv:2508.10104}, 2025.

\bibitem[Tajbakhsh et~al.(2015)Tajbakhsh, Gurudu, and Liang]{ColonDB}
Nima Tajbakhsh, Suryakanth~R Gurudu, and Jianming Liang.
\newblock Automated polyp detection in colonoscopy videos using shape and
  context information.
\newblock \emph{IEEE transactions on medical imaging}, 35\penalty0
  (2):\penalty0 630--644, 2015.

\bibitem[Tang et~al.(2022)Tang, Yang, Li, Roth, Landman, Xu, Nath, and
  Hatamizadeh]{tang2022self}
Yucheng Tang, Dong Yang, Wenqi Li, Holger~R Roth, Bennett Landman, Daguang Xu,
  Vishwesh Nath, and Ali Hatamizadeh.
\newblock Self-supervised pre-training of swin transformers for 3d medical
  image analysis.
\newblock In \emph{CVPR}, pages 20730--20740, 2022.

\bibitem[Tian et~al.(2025)Tian, Liao, Huang, Yang, Lei, Ourselin, and
  Liu]{tian2025endomamba}
Qingyao Tian, Huai Liao, Xinyan Huang, Bingyu Yang, Dongdong Lei, Sebastien
  Ourselin, and Hongbin Liu.
\newblock Endomamba: an efficient foundation model for endoscopic videos via
  hierarchical pre-training.
\newblock In \emph{MICCAI}, pages 224--234. Springer, 2025.

\bibitem[Tian et~al.(2022)Tian, Pang, Liu, Liu, Wang, Chen, Verjans, and
  Carneiro]{polypdiag}
Yu~Tian, Guansong Pang, Fengbei Liu, Yuyuan Liu, Chong Wang, Yuanhong Chen,
  Johan Verjans, and Gustavo Carneiro.
\newblock Contrastive transformer-based multiple instance learning for weakly
  supervised polyp frame detection.
\newblock In \emph{MICCAI}, pages 88--98. Springer, 2022.

\bibitem[Tiu et~al.(2022)]{tiu2022expert}
Ekin Tiu et~al.
\newblock Expert-level detection of pathologies from unannotated chest x-ray
  images via self-supervised learning.
\newblock \emph{Nature Biomedical Engineering}, 6\penalty0 (12):\penalty0
  1399--1406, 2022.

\bibitem[V{\'a}zquez et~al.(2017)V{\'a}zquez, Bernal, S{\'a}nchez,
  Fern{\'a}ndez-Esparrach, L{\'o}pez, Romero, Drozdzal, and
  Courville]{EndoScene}
David V{\'a}zquez, Jorge Bernal, F~Javier S{\'a}nchez, Gloria
  Fern{\'a}ndez-Esparrach, Antonio~M L{\'o}pez, Adriana Romero, Michal
  Drozdzal, and Aaron Courville.
\newblock A benchmark for endoluminal scene segmentation of colonoscopy images.
\newblock \emph{Journal of healthcare engineering}, 2017\penalty0 (1):\penalty0
  4037190, 2017.

\bibitem[Wang et~al.(2025{\natexlab{a}})Wang, Tang, Yao, Yan, Zhang, Huang,
  Lai, He, Tao, Jiang, et~al.]{wang2025simcrop}
Rongsheng Wang, Fenghe Tang, Qingsong Yao, Rui Yan, Xu~Zhang, Zhen Huang,
  Haoran Lai, Zhiyang He, Xiaodong Tao, Zihang Jiang, et~al.
\newblock Simcrop: Radiograph representation learning with similarity-driven
  cross-granularity pre-training.
\newblock In \emph{MICCAI}, pages 563--573. Springer, 2025{\natexlab{a}}.

\bibitem[Wang et~al.(2024)Wang, Zhao, Marostica, Yuan, Jin, Zhang, Li, Tang,
  Wang, Li, et~al.]{wang2024pathology}
Xiyue Wang, Junhan Zhao, Eliana Marostica, Wei Yuan, Jietian Jin, Jiayu Zhang,
  Ruijiang Li, Hongping Tang, Kanran Wang, Yu~Li, et~al.
\newblock A pathology foundation model for cancer diagnosis and prognosis
  prediction.
\newblock \emph{Nature}, 634\penalty0 (8035):\penalty0 970--978, 2024.

\bibitem[Wang et~al.(2023)Wang, Liu, Zhang, and Dou]{endofm}
Zhao Wang, Chang Liu, Shaoting Zhang, and Qi~Dou.
\newblock Foundation model for endoscopy video analysis via large-scale
  self-supervised pre-train.
\newblock In \emph{MICCAI}, pages 101--111. Springer, 2023.

\bibitem[Wang et~al.(2025{\natexlab{b}})Wang, Liu, Zhu, Wang, Zhang, and
  Dou]{endofmlv}
Zhao Wang, Chang Liu, Lingting Zhu, Tongtong Wang, Shaoting Zhang, and Qi~Dou.
\newblock Improving foundation model for endoscopy video analysis via
  representation learning on long sequences.
\newblock \emph{IEEE Journal of Biomedical and Health Informatics},
  2025{\natexlab{b}}.

\bibitem[Wang et~al.(2022)Wang, Wu, Agarwal, and Sun]{wang2022medclip}
Zifeng Wang, Zhenbang Wu, Dinesh Agarwal, and Jimeng Sun.
\newblock Medclip: Contrastive learning from unpaired medical images and text.
\newblock In \emph{EMNLP}, pages 3876--3887, 2022.

\bibitem[Xu et~al.(2021)Xu, Zhao, Yu, Zhang, Bian, Wang, Ge, and
  Qian]{xu2021real}
Jianwei Xu, Ran Zhao, Yizhou Yu, Qingwei Zhang, Xianzhang Bian, Jun Wang,
  Zhizheng Ge, and Dahong Qian.
\newblock Real-time automatic polyp detection in colonoscopy using feature
  enhancement module and spatiotemporal similarity correlation unit.
\newblock \emph{Biomedical Signal Processing and Control}, 66:\penalty0 102503,
  2021.

\bibitem[Yang et~al.(2025)Yang, Li, Yang, Zhang, Hui, Zheng, Yu, Gao, Huang,
  Lv, et~al.]{yang2025qwen3}
An~Yang, Anfeng Li, Baosong Yang, Beichen Zhang, Binyuan Hui, Bo~Zheng, Bowen
  Yu, Chang Gao, Chengen Huang, Chenxu Lv, et~al.
\newblock Qwen3 technical report.
\newblock \emph{arXiv preprint arXiv:2505.09388}, 2025.

\bibitem[Zhang et~al.(2023)Zhang, Xu, Usuyama, Xu, Bagga, Tinn, Preston, Rao,
  Wei, Valluri, et~al.]{zhang2023biomedclip}
Sheng Zhang, Yanbo Xu, Naoto Usuyama, Hanwen Xu, Jaspreet Bagga, Robert Tinn,
  Sam Preston, Rajesh Rao, Mu~Wei, Naveen Valluri, et~al.
\newblock Biomedclip: a multimodal biomedical foundation model pretrained from
  fifteen million scientific image-text pairs.
\newblock \emph{arXiv preprint arXiv:2303.00915}, 2023.

\end{thebibliography}
